\documentclass[10pt,twocolumn,letterpaper]{article}

\usepackage{iccv}
\usepackage{times}
\usepackage{epsfig}
\usepackage{graphicx}
\usepackage{amsmath}
\usepackage{amssymb}
\usepackage{todonotes}
\usepackage{mathtools}
\usepackage{algorithm}
\usepackage{algpseudocode}

\newcommand{\mathbold}[1]{\ensuremath{\boldsymbol{#1}}}

\newcommand{\mbb}{\mathbold{b}}

\newcommand{\mbh}{\mathbold{h}}

\newcommand{\mbx}{\mathbold{x}}

\newcommand{\mbz}{\mathbold{z}}

\newcommand{\mbI}{\mathbold{I}}

\newcommand{\mbalpha}{\mathbold{\alpha}}
\newcommand{\mbbeta}{\mathbold{\beta}}
\newcommand{\mbdelta}{\mathbold{\delta}}
\newcommand{\mbepsilon}{\mathbold{\epsilon}}

\newcommand{\mbpi}{\mathbold{\pi}}

\newcommand{\mbTheta}{\mathbold{\Theta}}

\newcommand{\diag}{\textrm{diag}}

\DeclareMathOperator*{\argmin}{arg\,min}

\newcommand{\mnv}[1]{\text{MobileNetV{#1}}}
\newcommand{\efnet}[1]{\text{EfficientNet-{#1}}}

\DeclarePairedDelimiter\round{\lfloor}{\rceil}

\usepackage{booktabs}
\usepackage{multirow}
\usepackage{caption}
\usepackage{subcaption}
\usepackage{xspace}

\usepackage[pagebackref=true,breaklinks=true,letterpaper=true,colorlinks,bookmarks=false]{hyperref}

\iccvfinalcopy 

\def\iccvPaperID{10801} 

\ificcvfinal\fi

\newcommand{\bwt}{bitwidth\xspace}
\newcommand{\bwts}{bitwidths\xspace}
\newcommand\mpb[1]{#1$_\text{MP}$}

\hypersetup{
    colorlinks=true,
    linkcolor=blue,
    filecolor=magenta,      
    urlcolor=cyan}
\begin{document}

\title{QBitOpt: Fast and Accurate Bitwidth Reallocation during Training}

\author{
Jorn Peters, Marios Fournarakis, Markus Nagel, Mart van Baalen, Tijmen Blankevoort\\
Qualcomm AI Research\thanks{Qualcomm AI Research, an initiative
of Qualcomm Technologies, Inc.}\\
Amsterdam, The Netherlands\\
{\tt\small \{jpeters,mfournar,markusn,mart,tijmen\}@qti.qualcomm.com}
}

\maketitle
\ificcvfinal\thispagestyle{empty}\fi

\begin{abstract}
Quantizing neural networks is one of the most effective methods for achieving efficient inference on mobile and embedded devices. In particular, mixed precision quantized (MPQ) networks, whose layers can be quantized to different \bwts, achieve better task performance for the same resource constraint compared to networks with homogeneous \bwts. However, finding the optimal bitwidth allocation is a challenging problem as the search space grows exponentially with the number of layers in the network. In this paper, we propose QBitOpt, a novel algorithm for updating bitwidths during quantization-aware training (QAT).
We formulate the \bwt allocation problem as a constraint optimization problem. 
By combining fast-to-compute sensitivities with efficient solvers during QAT, 
QBitOpt can produce mixed-precision networks with high task performance guaranteed to satisfy strict resource constraints. This contrasts with existing mixed-precision methods that learn \bwts using gradients and cannot provide such guarantees. We evaluate QBitOpt on ImageNet and confirm that we outperform existing fixed and mixed-precision  methods under average \bwt constraints, commonly found in literature. 
\end{abstract}

\section{Introduction}
In the last decade, neural networks have driven some of the most significant advances in artificial intelligence. However, searching for better-performing neural networks has increased their resource requirements substantially. The increase in parameters and training time is most notable in transformer models that are becoming ubiquitous in computer vision~\cite{dosovitskiy2020ViT, touvron2022deit} and natural language understanding~\cite{radford2018GPT, brown2020gpt3}. As a result, deploying these ever-growing neural networks on resource-constrained devices, such as mobile phones, embedded systems, and IoT devices, remains a challenge. Quantization has been proven to be a very effective method of addressing this constraint by reducing the memory and computational cost of neural networks without sacrificing their accuracy~\cite{2014horowitz, Krishnamoorthi2018Whitepaper, Nagel2021Whitepaper}. This is achieved by compressing the floating-point weights and activation to a more efficient low \bwt fixed-point representation. Deploying quantizing neural networks on devices is also becoming easier as increasing numner of devices support efficient low-precision operation.

\begin{figure}[t]
\begin{center}
\includegraphics[width=0.95\linewidth]{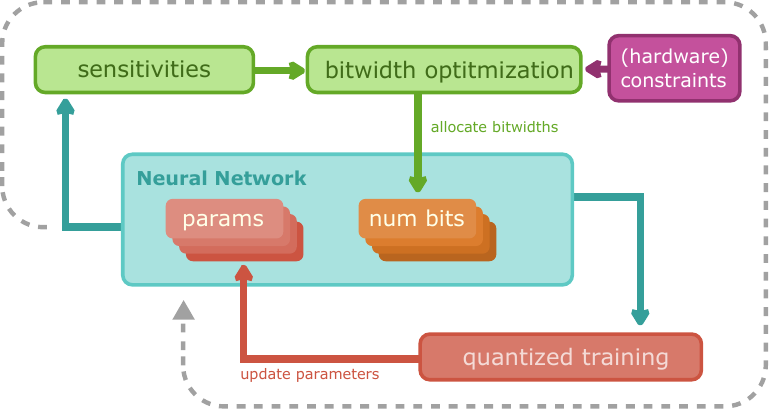}
\end{center}
\vspace{-.3cm}
   \caption{QBitOpt uses a quantization sensitivity metric to infer bitwidths while
   taking (hardware) constraints into account. The bitwidth allocation is used within a QAT training loop.}
\label{fig:hapqn_overview}
\end{figure}

Neural network quantization is an active field of research, and various methods have emerged in recent years. Broadly, we can distinguish them into two main groups: \textit{post-training quantization}~(PTQ)~\cite{nagel2019dfq, wang2020bitsplit,nagel2020adaround, frantar2022optimalBC, li2021brecq, frantar2022gptq}, and \textit{quantization-aware training}~(QAT)~\cite{jacob2018cvpr, Esser2020LSQ, Jain2019tqt, nagel2022oscillations, Uhlich2020DiffQ, Lee201EWGS}. In PTQ, we take a pre-trained neural network as input and infer the quantization setting that maximizes task performance without access to labeled data or the original training pipeline. In contrast, in QAT, we simulate quantization during training and allow the network parameters to adjust to quantization noise. In both cases, the target quantization bitwidth is commonly determined beforehand and is set homogeneously over the entire network. A crucial drawback of this approach is that the homogeneous bitwidth can only be as low as the network's most sensitive layer allows.

However, not all layers in a neural network are equally sensitive to quantization. Hence, it is beneficial to quantize some layers at a higher bitwidth than others. For example, it is common in quantization literature~\cite{Esser2020LSQ, Lee201EWGS, Defossez2021Diffq, nagel2022oscillations} to keep the first and last layer of a neural network at a higher bitwidth than the rest of the network because it improves accuracy with only an incremental increase in resource requirements. Allowing for a heterogeneous bitwidth allocation across the layers of a neural network is called \textit{mixed-precision quantization}~(MPQ). The main challenge of MPQ algorithms is to determine the optimal bitwidth per layer that achieves the highest task performance while minimizing resource requirements. In general, these are opposing objectives leading to a trade-off. What makes the MPQ particularly hard is that the search space grows exponentially in the number of neural network layers.

Since the \bwt allocation problem is a trade-off between neural network resources and task performance, there
will be many Pareto optimal solutions to the problem. However, in many real-world applications, the specific deployment target and associated resource constraints are often known beforehand. Assuming this, the MPQ \bwt allocation problem becomes a constrained single objective optimization problem. Leveraging this observation,  we propose QBitOpt, a novel method for inferring near-optimal \bwt allocations during QAT. We achieve this by measuring the \textit{sensitivity} of each layer to quantization noise and assigning them bitwidths that (1) satisfy the total resource requirements constraints and (2) minimize the total network quantization sensitivity. Our method stands out from previous mixed-precision work in the following ways:
\begin{itemize}
    \item It outputs a quantized neural network that is guaranteed to satisfy resource constraints. Most existing methods rely on hyperparameter search to balance the accuracy/resource constraints and can not guarantee the constraint is met. 
    \item By formulating the bitwidth allocation problem as a constrained convex optimization problem, our method scales  to networks that use many quantizers and can be solved quickly and efficiently using off-the-shelf software.
    \item We are the first to integrate optimization-based bitwidth allocation with existing quantization-aware training methods and outperform competing mix-precision methods in ImageNet classification under average \bwts constraints.
    \item We show that updating the bitwidth allocation during training is crucial for optimal performance and that it outperforms common post-training bitwidth allocation followed by quantization-aware fine-tuning.
\end{itemize}  

\section{Neural network quantization}
\label{sec:backgroud}
\label{sec:nn_quantizaiton}
\begin{figure}[t]
\begin{center}
\includegraphics[width=0.9\linewidth]{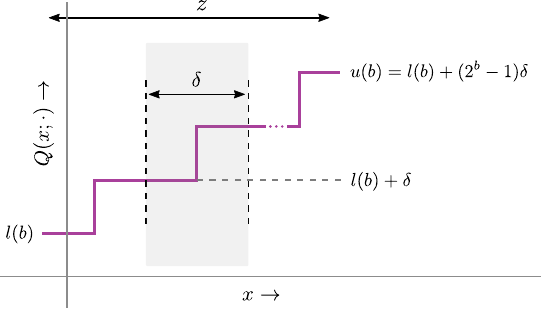}
\end{center}
\vspace{-.3cm}
   \caption{Schematic of the quantization operation of eq.~\eqref{eq:quantdef}}. 
\label{fig:quantizer_overview}
\end{figure}

The purpose of neural network quantization is to reduce the resource requirements and improve the latency of neural network inference while maintaining the task performance as much as possible and without changing the original network architecture. To achieve this we quantize the parameters and activations of a network to a quantization grid that is commonly learned during QAT ~\cite{Esser2020LSQ,choi2018pact,Uhlich2020DiffQ}.
In this paper, we define the quantization operation as follows:
\begin{align}
    Q(x; z, \delta, b) = \delta\cdot\mathrm{clamp}\left(\left\lfloor \frac{x + z}{\delta} \right\rceil;  l(b), u(b)\right) - z.
    \label{eq:quantdef}
\end{align}
Here $x$ denotes the quantizer input (\ie network parameters or activations), $z$ the zero-point of the quantizer,  and $\delta$ the quantizer step-size. $u(b)$ and $l(b)$ map the quantizer \bwt to the upper and lower clamping threshold in the integer domain. These can differ depending on the quantizer specifics (\eg (a)symmetric or (un)signed quantizers). $\round{\cdot}$ denotes the round-to-nearest integer mapping. See Figure~\ref{fig:quantizer_overview} for a graphical depiction of the quantizer defined in~\eqref{eq:quantdef}.

A neural network is quantized by a set of $K$ quantizer $\{Q_1, \ldots, Q_K\}$,
each associated with a different part of the neural network, such as parameters or activations.
In this paper, we assume \textit{symmetric uniform quantization}~\cite{Nagel2021Whitepaper} (\ie $z=0$) for all quantizers in the network. Moreover, for quantizer $i$, the step-size $\delta_i$ is defined in terms of the quantizer \bwt $b_i$ and a learned range parameter $\alpha_i$:
\begin{align}
    \delta_i = \frac{\alpha_i}{2^{b_i}-1}.
\end{align}
For ease of presentation, throughout this paper, we assume unsigned integer quantization.
For more details on neural network quantization, we refer the reader to~\cite{Nagel2021Whitepaper, gholami2021survey}.

To enable gradient-based learning during QAT, we need to overcome the fact that the gradient of the step function is zero almost everywhere. A common solution is to approximate the gradient of the quantization function \wrt its input with the straight-through estimator (STE)~\cite{hinton2012ste, bengio2013estimating}. \cite{Esser2020LSQ, Jain2019tqt} extend this approach to allow for gradient-based range learning.  

\subsection{Pseudo-quantization noise}
\label{sec:pseudo_quantization_noise}
Quantization has been studied in-depth in classical signal processing literature~\cite{gray1998quantization}. A common assumption is that the quantization error $\epsilon = x - Q(x)$ is uniformly distributed with zero-mean and a standard deviation of $\delta/\sqrt{12}$, where $\delta$ denotes the step-size of the quantizer $Q$. This assumption can be shown to hold exactly under some
conditions~\cite{widrow1956study,widrow1961statistical}. The relationship between quantization noise and quantization is studied in-depth in~\cite{Widrow1996-cv} and, when used in practice, the noise $\epsilon_q  = x - Q(x)\sim \mathcal{U}\left[-\frac{\delta}{2}, \frac{\delta}{2}\right]$ is referred to as \textit{pseudo-quantization noise}~(PQN).

Following the PQN formulation, we define the pseudo-quantization noise quantizer $Q_P$ as follows:
\begin{align}
    Q_P(x; z, \delta, b) = \delta \cdot \mathrm{clamp}\left(\frac{x + z}{\delta} + \epsilon; l(b), u(b) \right) - z.
    \label{eq:pqnquantdef}
\end{align}
Here $\epsilon \sim \mathcal[-\frac12, \frac12]$ and everything else remains the same as in~\eqref{eq:quantdef}. 
The PQN formulation is an attractive proposition for QAT because it is differentiable and, for this reason, it has been studied by~\cite{Defossez2021-jw,Park2022NIPQ}. However, there is no concrete evidence that it outperforms hard quantization with STE.

\subsection{Mixed-precision quantization objective}
\label{sec:mpqobjective}
In mixed-precision quantization, we want to solve a multi-objective optimization problem, which involves maximising the task performance while minimizing resource requirements.
Assuming a neural network with parameters $\mbTheta$, we quantize each part of this network is quantized by a separate quantizer $Q_i \in \{Q_1, \ldots, Q_K\}$. For example, one quantizer for parameters and activations per layer. Let $b_i \in \mbb = [b_1, \ldots, b_K]$ denote the \bwt associated with quantizer $Q_i$. We denote the quantized forward pass of this neural network, using a specific \bwt allocation as $Q(\mbTheta, \mbb)$ and the quantized task loss as $\mathcal{L}(Q(\mbTheta, \mbb))$ for an arbitrary task loss $\mathcal{L}$.

To complete the MPQ task, we introduce a resource cost $\pi(\mbb)$, \eg, memory requirements, CPU cycle counts, or run-time.
Throughout this paper, we assume the resource cost (or constraint) is either differentiable or a differentiable relaxation
exists. An often used approach to MPQ task is minimizing a scalarized objective~\cite{Baalen2020BayesianBits, Uhlich2020DiffQ, Park2022NIPQ} similar to:
\begin{align}
    \mathcal{L}_\textrm{MPQ}(\mbTheta, \mbb) =
    \lambda \mathcal{L}(Q(\mbTheta; \mbb)) + \pi(\mbb),
    \label{eq:basicmpqobj}
\end{align}
where $\lambda > 0$ is a scaling factor that weights the relative importance between the task loss and the resource cost. This approach is common because it allows us to learn the bitwidths and network parameters simultaneously, but it suffers from two main drawbacks: (1) it requires searching over values of $\lambda$ to find a trade-off that is acceptable to the user, and (2) there are no guarantees that the constraint will be met (cf. section~\ref{sec:literature_comparison}). Our proposed method is designed to mitigate these constraints. 
\label{sec:introduction}
\section{QBitOpt}
We introduce QBitOpt, our novel method for mixed-precision quantization (MPQ).
We previously established that the MPQ objective is a multi-objective optimization problem. 
As a result, the solution to the scalarized objective from equation~\eqref{eq:basicmpqobj} will depend on the weighting term $\lambda$. In practice, finding the optimal $\lambda$ between the two 
objectives that meet real-life resource constraints can be cumbersome and time-consuming. Instead, choosing a strict upper bound on the resources and optimizing the task performance within these bounds would be beneficial. 

We adopt this approach for QBitOpt, resulting in the following constrained optimization objective:
\begin{equation}
\begin{aligned}
    \textrm{minimize }\ \  & \mathcal{L}_\textrm{Q}(\mbTheta, \mbb) = \mathcal{L}(Q(\mbTheta, \mbb)) \\
    \textrm{subject to }\ \  & \widehat{\pi}(\mbb) \preceq 0
\end{aligned}    
\label{eq:consobj}
\end{equation}
 In the above, $\widehat{\pi}(\cdot)$ denotes the resource constraint and $\preceq$ element-wise inequality. This encodes a similar resource cost as $\pi(\cdot)$ in equation~\eqref{eq:basicmpqobj}. However, instead of minimizing the resource constraint, we only aim to satisfy an upper-bound constraint. Note that whereas we describe our method in terms of parameter quantization, this is not an intrinsic limitation of our method. We discuss details of activation quantization in appendix~\ref{appendix:actquant}.

Following the PQN model, we approximate the optimization objective further:
\begin{equation}
\begin{aligned}
    \mathcal{L}(Q_P(\mbTheta, \mbb)) &\approx \mathbb{E}_{\mbepsilon}\left[\mathcal{L}(\mbTheta + \mbepsilon')\right],
    \quad \mbepsilon' \sim \mathcal{U}[-\frac{\mbdelta}{2}, \frac{\mbdelta}{2}], \\
    &=\mathbb{E}_{\mbepsilon}\left[\mathcal{L}(\mbTheta + \frac{\mbalpha}{2^{b} - 1}\mbepsilon)\right],
    \quad \mbepsilon \sim \mathcal{U}[-\frac12, \frac12], \label{eq:pqnobjective}
\end{aligned}
\end{equation}
where $\mbdelta = [\delta_1, \ldots, \delta_{|\Theta|}]$ denotes the step-size associated with each parameter. Moreover, $\mbalpha$ and $\mbb$ \emph{broadcast} to the parameter dimensions. Following our formulation, the step size is a function of both the quantizer range $\mbalpha$ and the \bwt $\mbb$. The reparametrization in the second equality follows
from this. 
By splitting the optimization objective in~\eqref{eq:consobj} over the network parameters and the
bitwidth allocation into two separate minimizations, and substituting the result of equation~\eqref{eq:pqnobjective}, we obtain our QBitOpt optimization objective:
\begin{align}
    \min_{\mbTheta} \min_{\mbb} \mathbb{E}_{\mbepsilon}\left[  \mathcal{L}\left(\mbTheta + {\frac{\mbalpha}{2^{\mbb} - 1}}  {\mbepsilon} \right) \right] \quad \textrm{s.t. } \widehat{\pi}(\mbb) \preceq 0.
    \label{eq:splitobj}
\end{align}
This objective can now be optimized directly using, e.g., the penalty method~\cite{wright1999numerical} or the augmented Lagrangian method~\cite{wright1999numerical}. However, this would require
retraining the neural network to converge multiple times, which is costly. Instead, in QBitOpt, we efficiently solve an approximation to the inner minimization problem and use the
solution $\mbb^\ast$ to update $\mbTheta$ using common gradient-based QAT techniques.

\subsection{Approximate bitwidth minimization}
\label{sec:approx_bitwith_minimization}
The QBitOpt optimization, as stated in~\eqref{eq:splitobj}, can be solved directly, but doing so may be very costly. To enable gradient-based training
of the network parameters $\mbTheta$, we derive an efficient method for inferring the quantizer \bwt and solving the \bwt allocation problem. To this end,
we focus on the inner optimization problem of~\eqref{eq:splitobj}, that is,
\begin{align}
    \mbb^\ast = \argmin_{\mbb} \mathbb{E}_{\mbepsilon}\left[  \mathcal{L}\left(\mbTheta + {\frac{\mbalpha}{2^{\mbb} - 1}}  {\mbepsilon} \right) \right] \quad \textrm{s.t. } \widehat{\pi}(\mbb) \preceq 0.\label{eq:binneropt}
\end{align}
We first approximate~\eqref{eq:binneropt} using a second-order Taylor approximation w.r.t $\mbalpha/(2^{\mbb} - 1)$ around $\boldsymbol{0}$:
\begin{equation}
\begin{aligned}
    \mbb^\ast = \min_{\mbb} \mbh^\top \left(
    \frac{\mbalpha}{2^{\mbb}-1}
    \right)^2, \quad& \mbh_i = \nabla^2 \mathcal{L}(\mbTheta)_{ii} \\
    & \textrm{subject to } \widehat{\pi}(\mbb) \preceq 0,
    \label{eq:taylorobj}
\end{aligned}
\end{equation}
where $\mbh$ is the diagonal of the Hessian of the neural network. The first-order term cancels because the expectation of the pseudo-quantization noise is zero, and the constant term does not
depend on $\mbb$. For the full derivation, see appendix~\ref{appendix:ibm}.  The obtained objective is similar to the objective in~\cite{Dong2019Hawk}. However, they make an explicit assumption that the pre-trained floating-point network has converged, whereas we reach the same conclusion using only the zero-mean PQN assumption and make no assumptions about the state of the network. This distinction is important when solving~\eqref{eq:splitobj} because it allows us to solve this inner optimization during QAT regardless of whether the outer optimization over $\mbTheta$ has converged yet.

Depending on the exact choice of resource constraints $\widehat{\mbpi}$, \eqref{eq:taylorobj} is a convex program and can be solved very efficiently. In section~\ref{sec:choiceopt}, we discuss the choice of resource constraints and optimization methods in more detail.

Putting everything together, the QbitOpt optimization procedure consists of the following steps: first, solve~\eqref{eq:taylorobj} to obtain
$\mbb^\ast$, and second, perform backpropagation keeping $\mbb^\ast$ constant to update step for $\mbTheta$ and $\mbalpha$ using gradients: 
\begin{align}
    \nabla_{(\mbTheta, \mbalpha)} \mathbb{E}_{\mbepsilon}\left[ \mathcal{L}\left( \mbTheta + \frac{\mbalpha}{2^{\mbb^\ast}-1}\mbepsilon \right) \right]
\end{align}

\subsubsection{Quantization Sensitivities}
\label{sec:quant_sensitivities}
The optimization procedure we established for $\mbb$ in~\eqref{eq:taylorobj} depends on the diagonal of the Hessian of the neural network. In this optimization problem, the Hessian diagonal captures how sensitive a parameter or a group of parameters is to quantization or perturbations in general. We formally define \emph{sensitivity} as a statistic that quantifies how much the output of a neural network is affected by quantization. The diagonal of the Hessian in~\eqref{eq:taylorobj} is only one type of sensitivity, and one can imagine using other measures of sensitivity. In fact, this could be beneficial as calculating the Hessian diagonal of a neural network is often computationally expensive. This is why we use an approximation to the hessian diagonal called FIT~\cite{zandonati2022fit}. Another potential shortcoming of the Hessian is that it captures the effect of infinitesimal perturbations, whereas low-bit quantization perturbations may be pretty significant. We leave such an investigation for future work.

FIT is a sensitivity that has been shown to correlate well with the hessian diagonal~\cite{zandonati2022fit}. For this reason, we use FIT as a drop-in replacement for the
more expensive hessian diagonal. It is defined as:
\begin{align}
    \label{eq:fit}
    \widetilde{\mbh}_i = \left[\nabla_{\mbTheta} \mathcal{L}(\mbTheta)  \right]^2.
\end{align}
Note that FIT, similar to the hessian sensitivity, is computed based on the non-quantized neural network. As such, it requires an extra \emph{full precision} forward and
backward pass to compute and update sensitivities. Moreover, since we can only obtain gradients with respect to a subset of the training set, the computed FIT sensitivities
are stochastic in nature. To address the stochasticity in the sensitivity estimation and to reduce the computational requirements, we keep a running exponential moving average
of FIT sensitivities and only update them every $\tau$ iteration. See Section~\ref{sec:ema_abl} for more details on the choice of the EMA estimator.

So far, we have only focused on the quantization noise that stems from the rounding operator and can be effectively modeled as additive noise with PQN. However, the quantization operator of equation~\eqref{eq:quantdef} also involves \textit{clipping} of values outside the quantization grid limits. Clipping becomes an integral part of the training during QAT and can lead to considerable accuracy degradation if removed from the quantized model~\cite{alizadeh2020gradient}. For this reason, we compute FIT using the clipped model parameters $\widetilde{\mbTheta} = \textrm{clip}(\mbTheta, \mbalpha)$.

\subsection{Choice of Resource Constraints and Optimization Methods}
\label{sec:choiceopt}
So far, we have left the resource constraint $\widehat\pi(\cdot)$ undefined. In this section, we discuss the
choices for $\widehat\pi(\cdot)$ and its implications on the optimization problem. We aim to satisfy two goals:
First, the optimization problem must be efficiently solved to be integrated into training, and second, the constraints should express realistic resource constraints. We accomplish the first goal by restricting the constraints to be of the form:
\begin{align}
    \widehat{\pi}_i(\mbb) &\leq 0 \quad i = 1,\ldots,m \label{eq:inequality_constraint}\\ 
    \rho_j(\mbb) &= 0 \quad j = 1,\ldots,p, \label{eq:equality_constraint}
\end{align}
where the inequality constraint functions $ \widehat{\pi}_i(\mbb)$ are convex functions and the equality constraint functions $\rho_i(\mbb)$ are affine. Combining constraints of this form with the optimization objective~\eqref{eq:taylorobj}
results in a convex program\footnote{we referred to a single constraint $\widehat{\pi}$ up to this point. However, the convex program defined here allows for multiple inequality and equality constraints at the same time. Hence, $\widehat{\pi}$ refers to a set of constraints.}~\cite{boyd2004convex}. Such convex programs can be solved efficiently using off-the-shelf convex optimization libraries, such as~\cite{diamond2016cvxpy}, allowing us to solve the \bwt allocation problem frequently during training. In section~\ref{sec:ablation_studies}, we demonstrate that updating bitwidths during training leads to higher task performance. 

The convexity constraints of~\eqref{eq:inequality_constraint} and ~\eqref{eq:equality_constraint} may appear quite restrictive at first, but a lot of real-life constraints can be expressed this way, such as the \textit{BOPs}, the network's \textit{average bitwidth} or \textit{the network size bits}. Even if the resource constraints are not strictly convex, there is extensive literature on expressing non-convex problems as convex optimization problems~\cite[Part II]{boyd2004convex}.

In our constraint optimization objective so far, we have not restricted the solution $\mbb^\ast$ to be a positive integer, which is necessary for finding realistic solutions. Including such a constraint turns the optimization problem into an integer program which is, in general, NP-complete~\cite{papadimitriou1981complexity} and would be infeasible to solve for large networks. However, certain resource constraints admit efficient and specific optimization procedures. For example, the average \bwt constraint admits the \textit{greedy integer} method presented in Algorithm~\ref{alg:greedyint}. Alternatively, we can relax the integer constraint by allowing \textit{fractional} bits and solve the optimization problem efficiently using a convex program. We can use this relaxation during the \bwt update phase of training, where speed is important, and perform a final expensive integer program to infer the final \bwt allocation before the final fine-tuning stage. We discuss these choices in more detail in ablation studies of section~\ref{sec:ablation_studies}.

\subsection{Integrated QBitOpt QAT pipeline}
\label{sec:QAT_with_QBitOpt}
Algorithm~\ref{alg:qbitopt} describes our integrated QAT and QBitOpt pipeline. At each training iteration, we update the quantization sensitivities  $S_q$ for each quantizer using an exponential moving average, as described in~\eqref{sec:quant_sensitivities}, and perform a back-propagation step keeping the \bwt $\mbb$ to update the network $\mbTheta$ and quantization parameters $\mbalpha$. For every $\tau$ iteration, we update the quantizer \bwt, solving the convex optimization problem under the specified resource constraint, described in section~\ref{sec:choiceopt}. We do not update the bitwidths at every iteration in order to reduce the computational overhead and to give the network parameters enough time to adjust to the new \bwt allocation. For a high-level schematic, see Figure~\ref{fig:hapqn_overview}.


\begin{algorithm}
\caption{QBitOpt}\label{alg:qbitopt}
\begin{algorithmic}
\Require Neural network $\mathcal{N}$, task loss $\mathcal{L}(\cdot)$, resource constraints $\widehat{\pi}(\cdot)$, and sensitivity method $\mathcal{S}$
\For{iteration $i = 1, \ldots, T$}
    \For{quantizer $q = 1,\ldots,K$}
    \State $S^{(i)}_q = \gamma \mathcal{S}(\mathcal{L}(\mbTheta^{(i-1)}))_q + (1-\gamma)S^{(i-1)}_q$
    \EndFor
    \If{$i \textrm{ mod } \tau$ == 0}
        \State $\mbb = \textrm{BitOptimizer}(S_1^{(i)}, \ldots, S_k^{(i)})$
    \EndIf
    \State $\mbTheta^{(i)} = \mbTheta^{(i-1)} + \eta\nabla_{\mbTheta} \mathcal{L}(Q(\mbTheta^{(i-1)}; \mbb, \mbalpha^{(i-1}))$
    \State $\mbalpha^{(i)} = \mbalpha^{(i-1)} + \eta\nabla_{\mbalpha} \mathcal{L}(Q(\mbTheta^{(i-1)}; \mbb, \mbalpha^{(i-1}))$
\EndFor
\end{algorithmic}
\end{algorithm}

\begin{algorithm}
\caption{Greedy integer assignment for average \bwt}\label{alg:greedyint}
\begin{algorithmic}
\State \textbf{Initialize} $\mbb \leftarrow [1, \ldots, 1]$
\State $\mathcal{B} \leftarrow |\mbb| \cdot (\mbbeta-1)$
\State $\mathcal{S}_q \leftarrow $ sum of param./act. sensitivities for quantizer $q$
\For{iteration $1, \ldots, \mathcal{B}$}
    \State $q \leftarrow \argmin_j \mathcal{S}_j \left[ \mbalpha_j / (2^{\mbb_j} - 1) \right] $
    \State $\mbb_q \leftarrow \mbb_q + 1$
\EndFor
\end{algorithmic}
\end{algorithm}\label{sec:qbitopt}
\section{Related Work}







Neural network quantization has been a topic of interest in the research community for several years due to attractive hardware properties of quantized networks~\cite{courbariaux2014training,gupta2015deep,jacob2018cvpr,banner2018scalable,Esser2020LSQ}.
For recent in-depth surveys on PTQ and QAT topics, we refer the reader to \cite{Nagel2021Whitepaper} and \cite{gholami2021survey}.

Several methods for MPQ have been introduced in recent years. 
In gradient-based MPQ approaches~\cite{Baalen2020BayesianBits,Uhlich2020DiffQ,habi2020hmq,yang2021bsq}, a \bwt for each quantizer is learned through gradient-based optimization. An extension of gradient-based methods can be found in PQN-based approaches~\cite{Defossez2021-jw, Park2022NIPQ}. Using PQN instead of simulated quantization avoids the (biased) STE~\cite{hinton2012ste} and allows direct learning of each quantizer's \bwt. While these methods are efficient to run, defining a specific resource or accuracy target is impossible. Instead, a search over some regularization hyper-parameter is necessary. 
Several reinforcement learning techniques have been introduced~\cite{wang2019haq,rusci2020leveraging}, in which an agent selects quantization policies that balance hardware constraints with accuracy targets. These methods show promising results but require very long training times to achieve good performance.

Our work extends to a different branch of MPQ literature, in which \emph{sensitivity} is used as a statistic to select a \bwt for each quantizer. 
In these methods, quantizers with high quantization sensitivity are assigned higher \bwts.
Similarly to our work, the HAWQ line of research~\cite{Dong2019Hawk,dong2020hawq, yao2021hawqv3} defines sensitivity as either the spectral norm or the trace of the Hessian. 
Their approach significantly differs from ours in how sensitivity is incorporated:
due to the large computational requirements of their method, bitwidths are only estimated once, after which the network is iteratively fine-tuned.
\cite{zhao2021distribution} use a first-order Taylor expansion information to estimate sensitivity during QAT and reduce bitwidths for quantizers for the least sensitive quantizers. 
Since their method only allows iterative bitwidth reduction, it can never recover from faulty bitwidth assignments.
Taking a slightly different approach, \cite{pandey2023practical} measure a quantizer's sensitivity through signal-to-noise ratio on the network's output by lowering the bitwidth of a target quantizer. They then use this information to iteratively reduce each quantizer's bitwidth until some target accuracy or efficiency metric is reached. While the authors show good results, similarly to the HAWQ line of work, this approach is too slow to incorporate in QAT and is only used in PTQ settings.
FIT~\cite{zandonati2022fit} proposes a new sensitivity metric based on the trace of the Empirical Fisher matrix, which efficiently approximates the computationally expensive Hessian sensitivity in HAWQ. We use this sensitivity metric in QBitOpt to perform \bwt allocation during QAT efficiently.
Lastly, \cite{schaefer2023mixed} compares several sensitivity metrics for post-training MPQ.




\label{sec:relatedwork}
\section{Experiments}
In this section, we evaluate the effectiveness of QBitOpt by comparing it with other fixed-precision QAT methods and mixed-precision methods from the literature on the ImageNet \cite{imagenet} classification benchmark. We focus on low-bit quantization (4 and 3 bits on average) of efficient networks with depth-wise separable methods that are generally harder to quantize than fully convolutional networks \cite{nagel2022oscillations, Nagel2021Whitepaper}. 

\subsection{Experimental setup}
\label{sec:experimental_setup}
\paragraph{Quantization}
We follow the example of existing QAT literature and quantize the input to all layers except for the first layer and normalizing layers. In contrast to most existing QAT literature, we quantize \textit{all} layers, including the first and last layer, and let QBitOpt decide the optimal bitwidth for these layers. We quantize weights (per-channel) and activation (per-tensor) using hard quantization and train all network parameters, including the quantization threshold $\alpha$, using the straight-through estimator, similar to LSQ ~\cite{Esser2020LSQ}. 

\paragraph{Mixed-Precision}
Our training consists of two phases. In the first phase of \textit{mixed-precision} QAT, we calculate FIT sensitivities while training the quantized network and re-allocate bitwidths every $\tau$=250 training iterations. In the second (or \textit{fine-tuning}) phase, we freeze the obtained bitwidth allocation and fine-tune the remainder of the trainable parameters. This two-phase approach is quite common in the mixed-precision literature~\cite{Park2022NIPQ, Baalen2020BayesianBits}. Both phases are balanced in our experiments, and each takes up 50\%  of the training time unless stated otherwise. We study the effect of this choice in table~\ref{tbl:ablation_train_schedule_results}.

\paragraph{Resource Constraint}
In most experiments, we use the \textit{average bitwidth} constraint across all quantizers in the neural network, including activations and weights. We restrict the bitwidths to integers that are greater or equal to 2 bits and solve the optimization problem with the greedy algorithm~\ref{alg:qbitopt}. The abbreviation \mpb{$X$/$X$} below reflects this constraint with a target average bitwidth of $X$. 

In table~\ref{tbl:imagenet_results_weighted_avg_bits}, we additionally use the \textit{per-element average bitwidth} constraint, which is applied independently for weights and activations:
\begin{align}
    \rho_w(\mbb_w) &= \frac{\sum_i b_i^w \cdot e_i^w}{\sum_i e_i^w}, \quad \mbb_w\preceq 2, \label{eq:weight_constraint}\\
    \rho_a(\mbb_a) &= \frac{\sum_i b_i^a \cdot e_i^a}{\sum_i e_i^a}, \quad \mbb_a\preceq 2, \label{eq:act_constraint}
\end{align} 
where $b_i^w$/$b_i^a$ and $e_i^w$/$e_i^a$ denote the \bwt and number of elements of the $i$\textsuperscript{th} layer's weight/activation, respectively. We include these results for a fair comparison to the NIPQ~\cite{Park2022NIPQ} method, which uses this constraint.
 
\paragraph{Optimization}
In all cases, we start from a pre-trained full-precision network (see appendix~\ref{app:checkpoints} for the origin of the checkpoints) and instantiate the weight and activation quantization parameters using MSE range estimation~\cite{Nagel2021Whitepaper}. We train \mnv{2} and \efnet{Lite} for 30 epochs with SGD and momentum of 0.9 and \mnv{3}-Small for 40 epochs. More details about the optimization can be found in the appendix \ref{app:optimization}.

\subsection{Ablation studies}\label{sec:ablation_studies}
As we discussed in section \ref{sec:qbitopt}, our QBitOpt framework can be easily combined with existing QAT methods and deal with fractional or integers \bwts thanks to its speed and flexibility. In this section, we explore how different quantization and optimization options influence the performance of our framework. 

\paragraph{Sensitivities computation}
\label{par:calculating_sensitivies}
\begin{table}[t]
\centering
\begin{tabular}{l|c|cc}
\toprule
Architecture      & W/A  & $\widetilde{\mbh}$\textsubscript{FP} & $\widetilde{\mbh}$\textsubscript{clip}  \\
\midrule
\mnv{2}       & \mpb{4/4}    &     69.44      &   \textbf{69.71}  \\
\midrule
\multirow[c]{2}{*}{\mnv{3}-Small} & \mpb{4/4}   &  {63.47}   &  \textbf{64.23} \\
                                &   \mpb{3/3}   &  {56.64}   &  \textbf{57.14} \\
\bottomrule
\end{tabular}
 \caption{QBitOpt using  clipped sensitivities $\widetilde{\mbh}$\textsubscript{clip} and  full-precision sensitivities $\widetilde{\mbh}$\textsubscript{FP}. Validation accuracy (\%) on ImageNet.}
 \label{tbl:ablation_clipped_sense}
\end{table}
In section~\ref{sec:quant_sensitivities}, we argued that \textit{clipped} network sensitivities are better suited for inferring bitwidth during QAT, compared to vanilla full-precision sensitivities, as PQN models the noise from the rounding operator but not any clipping. The results in table~\ref{tbl:ablation_clipped_sense} confirm our hypothesis and show that in all cases, using the clipped sensitivities leads to better accuracy. In some cases, such as for \mnv{3}, not clipping during sensitivity computation can result in a substantial accuracy drop. 

\begin{figure}[t]
     \centering
     \begin{subfigure}[b]{0.48\textwidth}
         \centering
         \includegraphics[width=.95\textwidth]{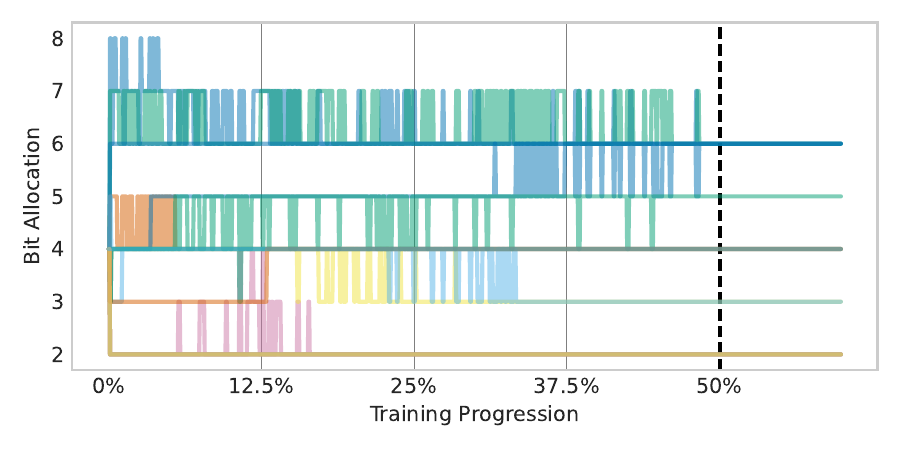}
\vspace{-.3cm}
         \caption{Bit allocation using \textit{greedy integer} bit optimization}
         \label{fig:greedy_integer}
     \end{subfigure}
     \hfill
     \begin{subfigure}[b]{0.48\textwidth}
         \centering
         \includegraphics[width=.95\textwidth]{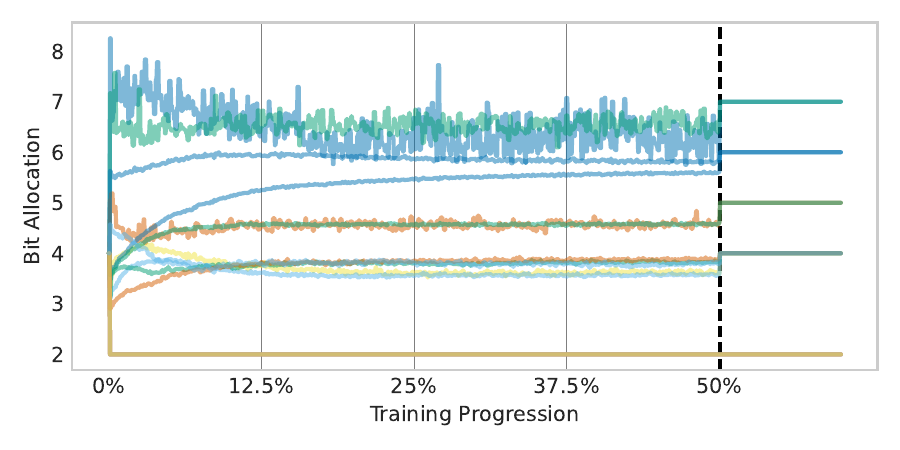}
\vspace{-.3cm}
         \caption{Bit allocation using fractional bit optimization}
         \label{fig:fractional}
     \end{subfigure}
        \caption{Bit allocation during training is reasonably stable, \ie, there are changes over time, but no sudden changes}
        \label{fig:bit_optimization}
\end{figure}

\paragraph{Fractional vs. integer \bwt}
\label{par:fractional_interger_bitwidth}
\begin{table}[t]
\centering
\begin{tabular}{l|c|cc}
\toprule
Architecture      & W/A    &    Fractional & Integer  \\ \midrule
\multirow[c]{2}{*}{\mnv{2}}       & \mpb{4/4}  &    67.91     & 69.71       \\
                                &  \mpb{3/3}   &   65.65      & 65.65      \\ \midrule
\multirow[c]{2}{*}{\mnv{3}-Small} & \mpb{4/4} &   64.23  & 64.39    \\
                                  & \mpb{3/3} &   57.14  &  57.14       \\
\bottomrule
\end{tabular}
 \caption{QBitOpt optimization method: integer vs. fractional bits. Validation accuracy (\%) on ImageNet.}
 \label{tbl:ablation_frac_integer}
\end{table}
In section~\ref{sec:choiceopt}, we discussed that constraining all bitwidths to natural numbers turns the convex program into an integer program, which is NP-hard. Here, we experimentally validate and compare our two proposed solutions, the \emph{greedy integer} method and \emph{fractional \bwt}.
In table~\ref{tbl:ablation_frac_integer}, we show the effect of both optimization methods on the final accuracy. We observe that in most cases, these lead to very similar performance, and there is no clear winner, except for \mnv{2} with a 4 bits target where greedy integers perform better. 
In figure~\ref{fig:bit_optimization}, we illustrate the progression of the \bwt allocation during training. We observe that for greedy integer (cf.~\ref{fig:greedy_integer}), some \bwts oscillate between two values which could introduce noise in the network optimization. On the other hand, fractional \bwts (cf.~\ref{fig:fractional} progress much smoother during \emph{mixed precision} phase, however at the end of the phase, there is a jump once the \bwt gets rounded to a viable natural number for the \emph{fine-tuning} phase.
For the remainder of the experiments, we favor greedy integers over fractional bitwidths.

\paragraph{Pseudo-noise vs. hard quantization}
\label{par:pseudo_vs_hard_quant}
\begin{table}[t]
\centering
\begin{tabular}{l|c|cc}
\toprule
Architecture      & W/A    &    Hard Quant & PQN  \\ \midrule
\multirow[c]{2}{*}{\mnv{2}}       & \mpb{4/4}  &     \textbf{69.71}    & 69.25    \\
                                  & \mpb{3/3}  &    \textbf{65.65}    &  64.21   \\ \midrule
\multirow[c]{2}{*}{\mnv{3}-Small} & \mpb{4/4}  &  \textbf{64.23}  &  53.92  \\
                                  & \mpb{3/3}   &    \textbf{57.14}     &  14.61   \\
\bottomrule
\end{tabular}
 \caption{Comparison of training with hard quantization and STE compared to PQN for the mixed-precision phase of QAT. Validation accuracy (\%) on ImageNet.}
 \label{tbl:ablation_hard_vs_pqn}
\end{table}

In the \emph{mixed precision} phase, we jointly train the network weights leveraging QAT while updating the \bwt allocation. Since updating the sensitivities is independent of the QAT update, we can choose any quantization method for QAT. 
Two natural choices are pseudo-quantization noise (PQN), which we used to derive our sensitivity metric, and hard-quantization with STE, which is commonly used in most QAT literature~\cite{jacob2018cvpr, Krishnamoorthi2018Whitepaper,  Esser2020LSQ, nagel2022oscillations}.
In table~\ref{tbl:ablation_hard_vs_pqn}, we experimentally compare both choices. Hard quantization with STE outperforms PQN for all models tested. Particularly for \mnv{3}-Small, the performance gap between the methods is quite significant. We hypothesize that it is important for the network to early in training adapt to the exact quantization noise present during inference.

\paragraph{Two-phase mixed-precision training}
\label{par:train_fraction_ablation}
\addtolength{\tabcolsep}{-1pt}
\begin{table}[t]
\centering
\begin{tabular}{l|c|ccccc}
\toprule
Arch.  & W/A &  0\% &  25\% & 50\% & 75\% & 100\%  \\ \midrule
\multirow[c]{2}{*}{MNv2} & \mpb{4/4} & 68.63 & 69.55   &	69.71   &	 69.7	& 67.81 \\
                             & \mpb{3/3} & 62.71 & 65.83   & 65.65    &  65.56 & 62.47 \\ \midrule
\multirow[c]{2}{*}{MNv3} & \mpb{4/4} & 59.62 & 64.39  & 64.23     &	64.34 &62.66 \\
                                  & \mpb{3/3} & 43.70 & 57.49   & 57.14    &	56.9  & 54.94 \\ \bottomrule
\end{tabular}
 \caption{Comparing training \mnv{2} (MNv2) and \mnv{3}-Small (MNv3) with QBitOpt for different fractions (\%) of the total training time. 0\% corresponds to using QBitOpt on a subset of training data to allocate bitwidths and freezing that allocation during training. Validation accuracy (\%) on ImageNet.}
 \label{tbl:ablation_train_schedule_results}
\end{table}
\addtolength{\tabcolsep}{1pt}
In this section, we study the impact of our two-phased mixed-precision training. In table~\ref{tbl:ablation_train_schedule_results}, we show the results with various divisions of the training time between the \emph{mixed precision} phase and the \emph{fine-tuning} phase.
Most importantly, these results demonstrate that both phases are important for obtaining the best mixed-precision performance.
When only fine-tuning is used after a post-training \bwt allocation (0\% column), the performance significantly degrades compared to updating the \bwt during training using QBitOpt --- especially for 3-bit quantization. This shows that our novel QBitOps algorithm, which integrates \bwt allocation into QAT, compares favorably to common post-training MPQ followed by fine-tuning.
On the other hand, updating the \bwt allocations up to the conclusion of training (100\% column) adversely affects performance. The non-constant bitwidths add extra noise to the optimization procedure. This can be harmful in the last phase of training, as discussed in the previous paragraph, and is a common issue known in the MPQ literature. As such, most training-based approaches circumvent this by not updating the \bwts towards the end of training~\cite{Baalen2020BayesianBits, Park2022NIPQ}.

While having both phases is clearly essential in finding the best mixed-precision network, our framework is relatively stable to the exact division of the training time between them. For most networks and \bwt, using between 25\%  and 75\%  for the mixed precision phase leads to good results.

\subsection{Comparison to existing methods}
\label{sec:literature_comparison}
\begin{table}[t]
    \centering
    \begin{tabular}{ c | l | c | c | c }
    \toprule
 &   Method & W/A & Avg. Bits & Acc. (\%)    \\
\midrule
\multirow[c]{10}{*}{\rotatebox[origin=c]{90}{\mnv{2}}}  
 & Full-precision & 32/32 & - & 71.72  \\
\cmidrule{2-5}
 & LSQ \cite{Esser2020LSQ}      & 4/4 & 4.0 & 69.45\textsuperscript{0.07}  \\
 & DQ \cite{Uhlich2020DiffQ}    & \mpb{4/4} & 4.08\textsuperscript{0.01} & 68.55\textsuperscript{0.30}  \\
& NIPQ \cite{Park2022NIPQ}    & \mpb{4/4} & 4.11\textsuperscript{0.0}  &    69.13\textsuperscript{0.07} \\
 & QBitOpt (ours)               &  \mpb{4/4} & \textbf{4.0} & \textbf{69.75}\textsuperscript{0.04}   \\
\cmidrule{2-5}
 & LSQ \cite{Esser2020LSQ}      & 3/3 & 3.0 & 65.17\textsuperscript{0.12}   \\
  & DQ \cite{Uhlich2020DiffQ}   & \mpb{3/3} & 3.10\textsuperscript{0.01} & 64.86\textsuperscript{0.07}  \\
  & NIPQ \cite{Park2022NIPQ}\textsuperscript{\dag}    & \mpb{3/3} & 3.15\textsuperscript{0.0} & 62.25\textsuperscript{0.5} \\
 & QBitOpt (ours)               &  \mpb{3/3} & \textbf{3.0} & \textbf{65.71}\textsuperscript{0.07}  \\
\midrule \midrule
\multirow[c]{10}{*}{\rotatebox[origin=c]{90}{\mnv{3}-Small} }
& Full-precision & 32/32 & - & 67.67  \\
\cmidrule{2-5}
& LSQ \cite{Esser2020LSQ}       & 4/4 & 4.0 & 61.58\textsuperscript{0.07} \\
& DQ \cite{Uhlich2020DiffQ}     & \mpb{4/4} & 4.10\textsuperscript{0.01} & 63.60\textsuperscript{0.02}  \\
& NIPQ \cite{Park2022NIPQ}      & \mpb{4/4} & 4.11\textsuperscript{0.01} & 62.21\textsuperscript{0.15} \\
 & QBitOpt (ours)               & \mpb{4/4}   & \textbf{4.0} & \textbf{64.34}\textsuperscript{0.06}  \\
\cmidrule{2-5}
  & LSQ \cite{Esser2020LSQ}     & 3/3 & 3.0 & 51.92\textsuperscript{0.04} \\
& DQ \cite{Uhlich2020DiffQ}     & \mpb{3/3} & 3.09\textsuperscript{0.01} & 57.26\textsuperscript{0.44}  \\
& NIPQ \cite{Park2022NIPQ} \textsuperscript{\dag}   & \mpb{3/3} & 3.22\textsuperscript{0.01} & 56.25\textsuperscript{0.72} \\
& QBitOpt (ours)                & \mpb{3/3}  & \textbf{3.0} & \textbf{57.36}\textsuperscript{0.17}  \\
\midrule \midrule
\multirow[c]{10}{*}{\rotatebox[origin=c]{90}{\efnet{lite}}}
& Full-precision & 32/32 & - & 75.42  \\
\cmidrule{2-5}
 & LSQ \cite{Esser2020LSQ}      & 4/4 & 4.0 & 72.93\textsuperscript{0.09}  \\
 & DQ \cite{Uhlich2020DiffQ}   & \mpb{4/4} & 4.08\textsuperscript{0.01} & 72.29\textsuperscript{0.21} \\
 & NIPQ \cite{Park2022NIPQ}      & \mpb{4/4} & 4.10\textsuperscript{0.01} &  72.27\textsuperscript{0.05}\\
 & QBitOpt (ours)              & \mpb{4/4} & \textbf{4.0} & \textbf{73.32}\textsuperscript{0.10}  \\
\cmidrule{2-5}
 & LSQ \cite{Esser2020LSQ}      & 3/3 & 3.0 & 69.62\textsuperscript{0.15}  \\
 & DQ \cite{Uhlich2020DiffQ}    & \mpb{3/3} & 3.13\textsuperscript{0.01} & 68.82\textsuperscript{0.21}  \\
 & NIPQ \cite{Park2022NIPQ} & \mpb{3/3} & 3.13\textsuperscript{0.01} & 66.56\textsuperscript{0.12} \\
 & QBitOpt (ours)               & \mpb{3/3} & \textbf{3.0} & \textbf{70.00}\textsuperscript{0.07}  \\
\bottomrule
\end{tabular}
\vspace{0.2em}
 \caption{Comparison of QBitOpt with other methods on the ImageNet image classification benchmark. We report validation accuracy (\%), target \bwt (W/A) set for the corresponding method and the achieved average \bwt (Avg. Bits) for mixed-precision methods. We run 3 seeds for all experiments and report the mean result and STD in the superscript.\textsuperscript{\dag} only 2 seeds available due to unstable training. 
 }
 \label{tbl:imagenet_results_avg_bits}
\end{table}

\begin{table}[t]
    \centering
    \begin{tabular}{ c | l | c | c }
    \toprule
 &   Method & Bits(W/A) & Acc. (\%)  \\
\midrule
\multirow[c]{4}{*}{\rotatebox[origin=c]{90}{MNv2}}
& NIPQ   & 4.08\textsuperscript{0.0}/4.09\textsuperscript{0.03} & \textbf{70.70}\textsuperscript{0.02}  \\
 & QBitOpt (ours)  &  4/4 & 70.39\textsuperscript{0.05} \\
\cmidrule{2-4}
  & NIPQ   & 3.1\textsuperscript{0.0}/ 3.08\textsuperscript{0.02} & 67.06\textsuperscript{0.28} \\
 & QBitOpt (ours)               & 3/3  & \textbf{68.44}\textsuperscript{0.0}  \\
\midrule \midrule
\multirow[c]{4}{*}{\rotatebox[origin=c]{90}{MNv3-Small}}
& NIPQ      & 4.209\textsuperscript{0.06}/4.154\textsuperscript{0.06}  & 61.01\textsuperscript{1.64} \\
 & QBitOpt (ours)               & 4/4  & \textbf{62.45}\textsuperscript{0.21}  \\
\cmidrule{2-4}
& NIPQ   & - & -  \\
& QBitOpt (ours)                & 3/3 & 57.82\textsuperscript{0.29} \\
\midrule \midrule
\multirow[c]{4}{*}{\rotatebox[origin=c]{90}{EffNet}}

 & NIPQ\textsuperscript{\dag}     & 4.065\textsuperscript{0.0}/ 4.119\textsuperscript{0.00} &  \textbf{73.62}\textsuperscript{0.03} \\
 & QBitOpt (ours)              & 4/4 & 73.01\textsuperscript{0.11}   \\
\cmidrule{2-4}
 & NIPQ   & 3.066\textsuperscript{0.03}/  3.089\textsuperscript{0.01}&  70.70\textsuperscript{0.14} \\
 & QBitOpt (ours)             &  3/3 &  \textbf{72.13}\textsuperscript{0.08} \\
\bottomrule
\end{tabular}
\vspace{0.2em}
 \caption{Comparison of QBitOpt and NIPQ~\cite{Park2022NIPQ} for the \textit{per-element average bitwidth} objective independently optimized for weights and activation. We report validation accuracy (\%) and the achieved \textit{per-element average bitwidth} for weights and activations. We run 3 seeds for all experiments and report the mean result and STD in the superscript.\textsuperscript{\dag}~no results available due to unstable training.}
 \label{tbl:imagenet_results_weighted_avg_bits}
\end{table}

We compare QbitOpt to other QAT approaches: (a) LSQ~\cite{Esser2020LSQ} for fixed-precision quantization; (b) \textit{differentiable quantization} (DQ)~\cite{Uhlich2020DiffQ} and \textit{noise injection pseudo quantization} (NIPQ)~\cite{Park2022NIPQ} for mixed-precision. For a fair comparison, we have re-implemented DQ and NIPQ, adhering to their implementation details and modifying them appropriately to match our constraints and quantization assumptions. We performed an extensive hyperparameter search on the regularization strength $\lambda$ and learning rates. For further details, refer to appendix~\ref{app:exp_results}.

In table~\ref{tbl:imagenet_results_avg_bits}, we present the results for \mnv{2}, \mnv{3}-Small and \efnet{Lite} on the ImageNet classification benchmark. QbitOpt significantly outperforms the strong fixed-precision LSQ baseline in all cases. Especially for the challenging to quantize \mnv{3}-Small, the accuracy improves up to 5\%. We also outperform the mixed-precision method, NIPQ \& DQ, in terms of quantized accuracy, even though NIPQ and DQ achieve a higher average \bwt. These results showcase one of the biggest strengths of our method, namely that the resource constraint is met exactly. This contrasts with the other mixed-precision methods, where we struggle to meet the target constraint even after an extensive hyperparameter search.

In table~\ref{tbl:imagenet_results_avg_bits}, we compare QBitpt to NIPQ using the \textit{per-element average bitwidth} implemented in their work~(cf. equations~\eqref{eq:weight_constraint} \& \eqref{eq:act_constraint}) for a fairer comparison. Under this less stringent constraint, the accuracy gap between the two methods closes, and in two cases, \mnv{2} W4A4 and \efnet{Lite} W4A4, NIPQ outperforms QBitOpt. However, the trend is reversed for 3 bits and, specifically, for \mnv{3}-Small training was unstable, and we could not reach satisfactory accuracy. Despite closing the accuracy margin, QBitOpt is the only method guaranteeing the resource constraint is met in all cases without searching for regularization strength, thus, significantly reducing experimentation time. 

 
\label{sec:experiments}

\section{Conclusion}
In this work, we introduced QBitOpt, a novel algorithm for allocating \bwts under strict resource constraints. QBitOpt leverages techniques from convex optimization and QAT to obtain high-performing neural networks in the sense of resource requirements and task performance without the need to carefully balance task loss and resource cost through a cumbersome hyper-parameter. 
We justify using Hessian-based sensitivities, traditionally only used for fully converged neural networks, during training. Our greedy integer method and efficient FIT approximation allow us to regularly update the \bwt allocations during quantization-aware training.

We demonstrate the efficacy of QBitOpt on several architectures in which QBitOpt compares favorably to competing fixed-precision and mixed-precision approaches. We further examine various properties of our method through ablation studies. Specifically, we show that regularly reallocating bitwidths during training is crucial for optimal performance. We identify improved sensitivities that more closely capture quantization perturbation and the design of convex relaxations for hardware constraints as exciting avenues for future research.
\label{sec:conclusion}
{\small
\bibliographystyle{ieee_fullname}
\bibliography{ref}
}

\newpage
\clearpage
\appendix
\section{Activation quantization}
\label{appendix:actquant}
In section~\ref{sec:qbitopt}, we formulated the QBitOpt  optimization objective in terms of network parameters $\mbTheta$. However, we commonly also quantize the activations in neural network quantization. Fortunately, our method extends easily to activations too, as it is trivial to compute quantization sensitivities for activations.
As a reminder, in section~\ref{sec:approx_bitwith_minimization}, we defined the Hessian diagonal $\mbh$ as an example of quantization sensitivity. By computing the \textit{hessian sensitivities} with respect to activations, we can infer bitwidth allocation for activation as per equation~\ref{eq:taylorobj}. 

Considering an arbitrary neural network $F$ with $L$ layers as a composition of functions:
\begin{align}
    F(\mbx) = \left[f_L \circ f_{L-1} \circ \cdots \circ f_2 \circ f_1\right](\mbx),
\end{align}
we can compute the sensitivity of the $i$\textsuperscript{th} activation $\mbz_i = \left[f_i \circ \cdots \circ f_1\right](\mbx)$ by simply considering the sub-network:
\begin{align}
    F_{i+1}= \left[f_L \circ f_{L-1} \circ \cdots \circ f_{i+1}\right]
\end{align}
The \textit{hessian sensitivity} is now given by:
\begin{align}
     \mathbb{E}_{\mbx \sim \mathcal{D}} \left[
    \frac{\partial^2}{\partial \mbz_i^2} \mathcal{L}\left(F_{i+1}(\mbz)\right)\biggr\rvert_{\mbz = \mbz_{i}}  \right] 
\end{align}

The sensitivity can be computed over the full dataset $\mathcal{D}$ using an exponential moving average, as described in section~\ref{sec:quant_sensitivities}. 
In practice, one may consider applying separate constraints to the parameter and activation quantizer bitwidth. We leave this investigation for future work.

\section{Inner \bwt minimization}
\label{appendix:ibm}
For convenience, we restate~\eqref{eq:splitobj} here:
\begin{align}
    \min_{\mbTheta} \min_{\mbb} \mathbb{E}_{\mbepsilon}\left[  \mathcal{L}\left(\mbTheta + {\frac{\mbalpha}{2^{\mbb} - 1}}  {\mbepsilon} \right) \right] \quad \textrm{s.t. } \widehat{\pi}(\mbb) \preceq 0.
\end{align}
As presented in the main text(cf. section~\ref{sec:approx_bitwith_minimization}), we aim to solve an approximation to the inner minimization. The solution to this is then used to take a gradient step for the outer minimization. First, let $\mbdelta = \mbalpha/(2^{\mbb} - 1)$. Then we approximate the objective using a second-order Taylor approximation:
\begin{equation}
\begin{aligned}
    &\mathbb{E}_{\mbepsilon}\left[\mathcal{L}(\mbTheta + \mbdelta\mbepsilon)\right]  \\
    & \approx 
    \mathbb{E}_{\mbepsilon}\left[ \mathcal{L}(\mbTheta) + (\mbdelta\mbepsilon)^\top\nabla_{\mbTheta}\mathcal{L}(\mbTheta)
    + \frac12(\mbdelta\mbepsilon)^\top \nabla^2_{\mbTheta} \mathcal{L}(\mbTheta)(\mbdelta\mbepsilon) \right]
    \label{losstaylorapprox}
\end{aligned}
\end{equation}
The first term is constant with respect to $\mbb$ and the second term equals zero as $\mathbb{E}_{\mbepsilon}[\mbepsilon] = 0$. This leaves the following optimization objective:
\begin{equation}
    \begin{aligned}
        &\frac12 \mathbb{E}_{\mbepsilon}\left[(\mbdelta\mbepsilon)^\top \left[\nabla^2 \mathcal{L}(\mbTheta)\right](\mbdelta\mbepsilon) \right] \\
        &\ \ = \frac12\mathrm{Tr}\left\{\mathbb{E}_{\mbepsilon}[(\mbdelta\mbepsilon)^\top \left[\nabla^2 \mathcal{L}(\mbTheta)\right](\mbdelta\mbepsilon)\right]\} \\
        &\ \ \stackrel{(a)}{=} \frac12\mathrm{Tr}\left\{\mathbb{E}_{\mbepsilon}[(\mbdelta\mbepsilon)(\mbdelta\mbepsilon)^\top \right]\left[\nabla^2 \mathcal{L}(\mbTheta)\right]\} \\
        &\ \ = \frac12\mathrm{Tr}\left\{\diag(\mbdelta)\mathbb{E}_{\mbepsilon}[\mbepsilon\mbepsilon^\top \right]\diag(\mbdelta)\left[\nabla^2 \mathcal{L}(\mbTheta)\right]\} \\
        &\ \ \stackrel{(b)}{=} \frac{1}{24}\mathrm{Tr}\left\{\diag(\mbdelta)\diag(\mbdelta)\left[\nabla^2 \mathcal{L}(\mbTheta)\right]\right\} \\
        &\ \ = \frac{1}{24}\mathrm{Tr}\left\{\diag(\mbdelta)^2\left[\nabla^2 \mathcal{L}(\mbTheta)\right]\right\} \\
        &\ \ = \frac{1}{24}\sum_{i}^{|\mbTheta|} \left[\nabla^2 \mathcal{L}(\mbTheta)\right]_{ii} \mbdelta^2_{ii}.
    \end{aligned}
\end{equation}
Here, $(a)$ uses $\mathrm{Tr}(A^\top B) = \mathrm{Tr}(BA^\top)$ where $A$ and $B$ are two $m \times n$ real matrices and $(b)$ follows from $\mathbb{E}[\mbepsilon\mbepsilon^\top] = \mbI/12$.
Finally, dropping the multiplicative constants that do not affect the minimization problem and re-substituting $\mbdelta$, we obtain our
optimization problem:
\begin{equation}
\begin{aligned}
    \mbb^\ast = \min_{\mbb} \mbh^\top \left(
    \frac{\mbalpha}{2^{\mbb}-1}
    \right)^2, \quad& \mbh_i = \nabla^2 \mathcal{L}(\mbTheta)_{ii} \\
    & \textrm{subject to } \widehat{\pi}(\mbb) \preceq 0.
\end{aligned}
\end{equation}

\section{Experimental configuration}
\label{app:experimantal_configuration}
\begin{table*}[t]
\centering
\begin{tabular}{c | c | c c | c c  c c | c c }

\toprule
\multirow[c]{2}{*}{Model}  & \multirow[c]{2}{*}{W/A} & \multicolumn{2}{c|}{Epochs} & \multicolumn{4}{c}{SGD - Model parameters} &  \multicolumn{2}{|c}{Adam - Quant. parameters} \\
\rule{0pt}{10pt}

 & & Phase-1 & Phase-2 & LR & $\eta_{min}$ & Warmup & WD & LR & WD\\
\midrule
\multirow[c]{2}{*}{\mnv{2}} &  4/4 & 15 & 15 &  0.0033 & 3.3$\times 10^{-6}$  & - & 1.0$\times 10^{-5}$ &  1.0$\times 10^{-5}$ & $0.0$ \\
&  3/3 & 15 & 15 &  0.01 & 1.0$\times 10^{-5}$  & - & 1.0$\times 10^{-5}$ &  1.0$\times 10^{-5}$ & $0.0$ \\
\midrule
\efnet{} &  4/4 & 15 & 15 &  0.0033 & 3.3$\times 10^{-6}$  & - & 1.0$\times 10^{-5}$ &  1.0$\times 10^{-5}$ & $0.0$ \\
Lite &  3/3 & 15 & 15 &  0.01 & 1.0$\times 10^{-5}$  & - & 1.0$\times 10^{-5}$ &  1.0$\times 10^{-5}$ & $0.0$ \\
\midrule
\mnv{3}- &  4/4 & 20 & 20 &  0.07 & 7.0$\times 10^{-5}$  & 4 & 1.0$\times 10^{-5}$ &  1.0$\times 10^{-5}$ & $0.0$ \\
Small &  3/3 & 20 & 20 &  0.05 & 5.0$\times 10^{-5}$  & 4 & 0.0 &  1.0$\times 10^{-5}$ & $0.0$ \\
\bottomrule
\end{tabular}
\caption{Optimization configuration: all model parameters (except for the quantization parameters) are optimized using SGD with a momentum of 0.9 and a cosine annealing schedule for learning rate. We use a separate Adam optimizer with a constant learning rate for the quantization parameters. WD: weight decay; Warmup: number of epoch for linear learning rate warmup; $\eta_{min}$: final learning rate of cosine decay. }
\label{tbl:optimization_config}
\end{table*}

\subsection{Optimization}
\label{app:optimization}
All experiments are trained on NVIDIA GPUs using  Python 3.8.10, PyTorch v1.11, and Torchvision v0.12. We found that using separate optimizers for the model (SGD) and quantization parameters (Adam) leads to stabler training and higher accuracy across architectures and methods. The optimizer and learning rate schedules for both optimizers can be found in table~\ref{tbl:optimization_config}. \textit{Phase-1} refers to QAT with \bwt reallocation using QBitOpt, whereas in \textit{phase-2}, we fix the quantizers' bitwidth and fine-tune with QAT. At the end of each training epoch, we re-estimate the batch-normalization statistics using 50 batches of training data, as per~\cite{nagel2022oscillations}.

\subsection{QBitOpt configuration}
\label{app:QBitOptConfiguration}
\begin{figure}
\centering
\includegraphics[width=0.95\linewidth]{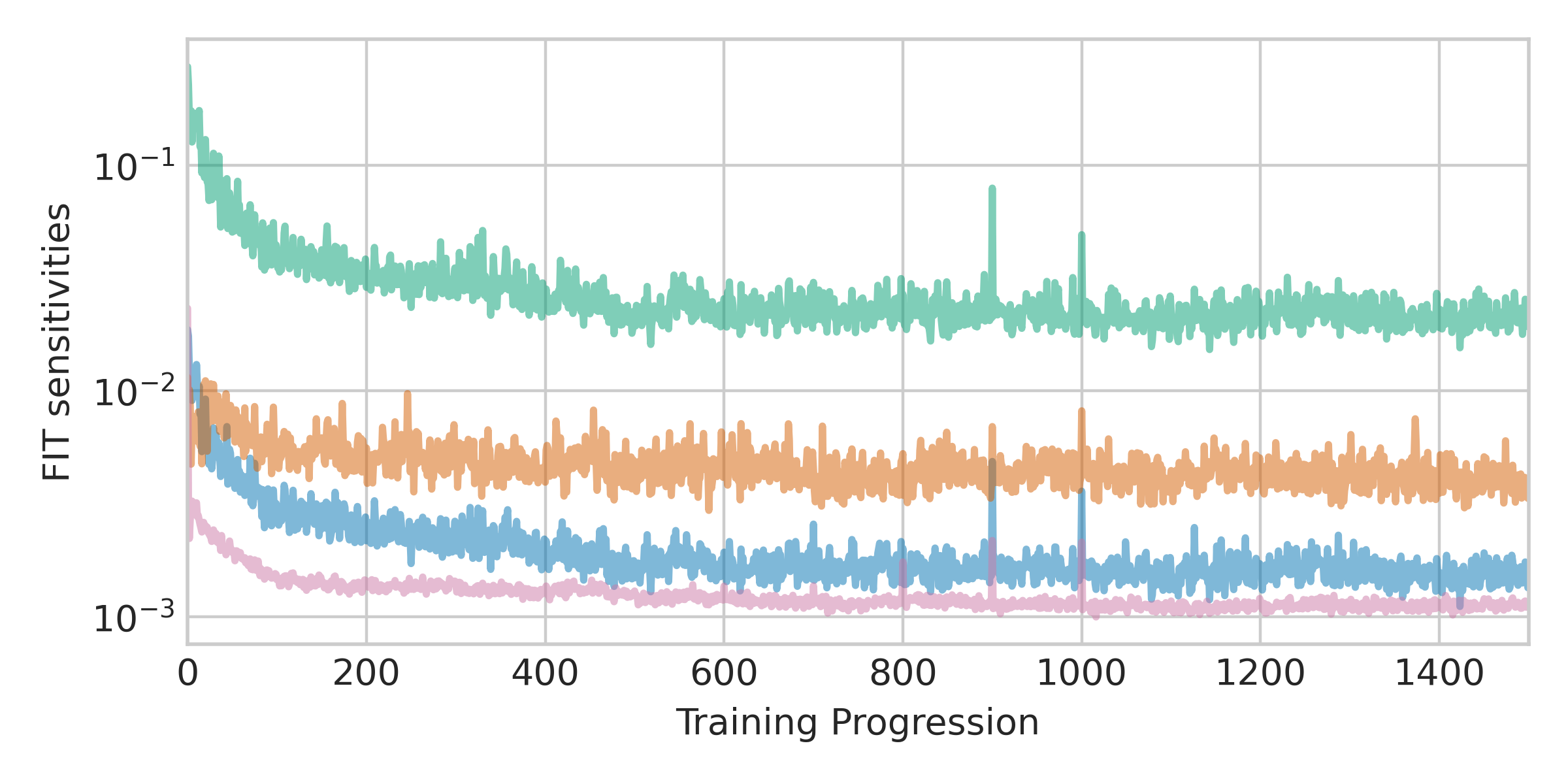}
 \vspace{-.3cm}
\caption{FIT sensitivities progression over time (y-axis in logarithmic scale)}
\label{fig:fit_sensitivities}
\end{figure}

During \textit{phase-1} of training, we calculate an exponential-moving average of each quantizer's FIT sensitivity $S_q$ with a momentum of 0.9, as shown in algorithm~\ref{alg:qbitopt}. Because sensitivities change quite smoothly during training (see fig.~\ref{fig:fit_sensitivities}), we decided to compute them only every two training iterations to reduce training time. We infer a new \bwt allocation by solving the optimization every $\tau=250$ training iterations. We found that QBitOpt is not very sensitive to this hyperparameter, and decreasing $\tau$ did not lead to better results. In fact, given that we learn the quantization range $\alpha$ using gradients, $\tau$ should be large enough to allow the quantization range to adapt to the latest bitwidth allocation. 

\subsection{Model checkpoints}
\label{app:checkpoints}
\begin{itemize}
    \item \mnv{2}: \url{https://github.com/tonylins/pytorch-mobilenet-v2} 
    
    \item \mnv{3}-Small: \url{https://pytorch.org/vision/0.12/models}
    \item \efnet{Lite}: \url{https://github.com/huggingface/pytorch-image-models}
\end{itemize}

\section{Additional experimental results}
\label{app:exp_results}
\subsection{Differential quantization (DQ)}
\label{app:dq_results}
\begin{table}[t]
\centering
\begin{tabular}{l r r}
\toprule

 Method & Avg. bits &  Acc. (\%)  \\
\midrule
DQ [$\lambda=0.3$] &	4.299	&64.38\\
DQ [$\lambda=0.5$]	&4.187	&63.87\\
DQ [$\lambda=0.7$]	&4.150	&63.92\\
DQ [$\lambda=1.0$]	&\textbf{4.093}	&\textbf{63.61}\\
DQ [$\lambda=1.4$]	&4.065	&63.56\\
DQ [$\lambda=2.0$]	&4.065	&63.42\\
DQ [$\lambda=3.0$]	&4.037 &63.44\\
DQ [$\lambda=5.0$]	&4.028	&63.44\\
DQ [$\lambda=8.0$]	& 4.009	&63.27\\
\midrule
QBitOpt &  \textbf{4.0} & \textbf{64.23} \\
\midrule
\midrule
DQ [$\lambda=0.3$] 	&3.542	&60.45\\
DQ [$\lambda=0.5$]	&3.402	&59.62\\
DQ [$\lambda=0.7$]	&3.290	&58.48\\
DQ [$\lambda=1.0$] &\textbf{3.215}	&\textbf{58.33}\\
DQ [$\lambda=1.4$]	&3.159	&57.95\\
DQ [$\lambda=2.0$]	&3.121	&57.63\\
DQ [$\lambda=3.0$]	&3.084	&56.86\\
DQ [$\lambda=5.0$]	&3.056	&55.70\\
DQ [$\lambda=8.0$]	&3.037	&56.89\\
\midrule
QBitOpt &\textbf{3.0} &  \textbf{57.14}\\
\bottomrule
\end{tabular}
\caption{\mnv{3}-Small on ImaneNet. DQ~\cite{Uhlich2020DiffQ} results for different regularization strength $\lambda$, including our QBitOpt result. The values in bold are used in table~\ref{tbl:imagenet_results_avg_bits}.}
\label{tbl:appedix_dq_mnv3}
\end{table}

\begin{table}[t]
\centering
\begin{tabular}{l r r}
\toprule

 Method & Avg. bits &  Acc. (\%)  \\
\midrule
DQ [$\lambda=0.3$]	&4.210	&69.09 \\
DQ [$\lambda=0.5$]	&4.143	&68.59\\
DQ [$\lambda=0.7$]	&4.105	&68.57\\
DQ [$\lambda=1.0$] &\textbf{4.076}	& \textbf{68.50}\\
DQ [$\lambda=1.4$]	&4.067	&68.51\\
DQ [$\lambda=2.0$]	&4.038	&68.80\\
\midrule
QBitOpt & \textbf{4.0} &  \textbf{69.71}\\
\midrule
\midrule
DQ [$\lambda=0.3$]	&3.286&	66.33\\
DQ [$\lambda=0.5$]	&3.229&	65.44\\
DQ [$\lambda=0.7$]	&3.124&	64.78\\
DQ [$\lambda=1.0$] & \textbf{3.114}& \textbf{64.94}\\
DQ [$\lambda=1.4$]	&3.086&	64.27\\
DQ [$\lambda=2.0$]	&3.048&	64.18\\
\midrule
QBitOpt & \textbf{3.0} &  \textbf{65.65}\\
\bottomrule
\end{tabular}
\caption{\mnv{2} on ImaneNet. DQ~\cite{Uhlich2020DiffQ} results for different regularization strength $\lambda$, including our QBitOpt result. The values in bold are used in table~\ref{tbl:imagenet_results_avg_bits}.}
\label{tbl:appedix_dq_mnv2}
\end{table}

\begin{table}[t]
\centering
\begin{tabular}{l r r}
\toprule

 Method & Avg. bits &  Acc. (\%)  \\
\midrule
DQ [$\lambda=0.3$] &4.242	&72.48\\
DQ [$\lambda=0.5$]	&4.141	&72.61\\
DQ [$\lambda=0.7$]	&4.111	&72.50\\
DQ [$\lambda=1.0$]&\textbf{4.081}	&\textbf{72.14}\\
DQ [$\lambda=1.4$]&4.061	&72.16\\
DQ [$\lambda=2.0$]&4.051	&72.20\\
\midrule
QBitOpt & \textbf{4.0 }& \textbf{ 73.43}\\
\midrule
\midrule
DQ [$\lambda=0.3$]	&3.354	&70.03\\
DQ [$\lambda=0.5$]	&3.212	&69.24\\
DQ [$\lambda=0.7$] &3.141	&68.95\\
DQ [$\lambda=1.0$]&\textbf{3.121}	&\textbf{68.86}\\
DQ [$\lambda=1.4$]&3.101	&68.48\\
DQ [$\lambda=2.0$]	&3.051	&68.33\\
\midrule
QBitOpt & \textbf{3.0} & \textbf{ 70.04}\\
\bottomrule
\end{tabular}
\caption{\efnet{Lite} on ImaneNet. DQ~\cite{Uhlich2020DiffQ} results for different regularization strength $\lambda$, including our QBitOpt result. The values in bold are used in table~\ref{tbl:imagenet_results_avg_bits}.}\label{tbl:appedix_dq_effnet}
\end{table}

In tables~\ref{tbl:appedix_dq_mnv2},~\ref{tbl:appedix_dq_mnv3} \&~\ref{tbl:appedix_dq_effnet}, we present the results of our hyperparameter search over the regularization strength $\lambda$ in the objective of \textit{differential quantization}  (DQ)~\cite{Uhlich2020DiffQ}. In contrast to QBitOpt, the method fails to reach the target average \bwt exactly. Notably, for \mnv{3}-Small, we have to extend the grid search significantly to achieve a satisfactory average \bwt solution. This study demonstrates the power of QBitOpt that does not really on a scalarized multi-objective loss. Instead, QBitOpt guarantees the exact constraint, enabling the user to focus on tuning the QAT hyperparameters and achieve the highest task performance under the specified constraint. Please note that the results in bold are shown in the final results  table~\ref{tbl:imagenet_results_avg_bits}.

\subsection{Noise injection pseudo quantization (NIPQ)}
\label{app:nipq_results}
\begin{table}[t]
\centering
\begin{tabular}{l r r}
\toprule

 Method & Avg. bits &  Acc. (\%)  \\
\midrule
NIPQ [$\lambda=2.0$]	& 4.229	& 69.39 \\
NIPQ [$\lambda=4.0$]	&   4.162 & 69.17\\
NIPQ [$\lambda=5.0$]	&	 4.140 & 69.16\\
NIPQ [$\lambda=8.0$] &\textbf{4.108}& \textbf{69.13}\\
NIPQ [$\lambda=9.0$]	&   4.111	&69.16\\
\midrule
QBitOpt & \textbf{4.0} &  \textbf{69.71}\\
\midrule
\midrule
NIPQ [$\lambda=3.0$]	& &	\\
NIPQ [$\lambda=4.0$]	& &	\\
NIPQ [$\lambda=5.0$]	& &	\\
NIPQ [$\lambda=7.0$]    & & \\
NIPQ [$\lambda=9.0$]	& &	\\
NIPQ [$\lambda=10.0$]	& &	\\
\midrule
QBitOpt & \textbf{3.0} &  \textbf{65.65}\\
\bottomrule
\end{tabular}
\caption{\mnv{2} on ImaneNet. NIPQ~\cite{Park2022NIPQ} results for different regularization strength $\lambda$, including our QBitOpt result. The values in bold are used in table~\ref{tbl:imagenet_results_avg_bits}.}
\label{tbl:appedix_NIPQ_mnv2}
\end{table}

\begin{table}[t]
\centering
\begin{tabular}{l r r}
\toprule

 Method & Avg. bits &  Acc. (\%)  \\
\midrule
NIPQ [$\lambda=3.0$] &4.192	&72.49\\
NIPQ [$\lambda=4.0$] & 4.182 &72.52\\
NIPQ [$\lambda=6.0$] & 4.131 &72.38\\
NIPQ [$\lambda=7.0$]& 4.125	&72.37\\
NIPQ [$\lambda=9.0$]& 	4.104 &72.32\\
NIPQ [$\lambda=10.0$]&	4.104 & 72.29 \\
NIPQ [$\lambda=11.0$]& \textbf{4.101}	& \textbf{72.29} \\
\midrule
QBitOpt & \textbf{4.0}& \textbf{73.43}\\
\midrule
\midrule
NIPQ [$\lambda=3.0$] &	3.283 & 68.49 \\
NIPQ [$\lambda=4.0$] & 3.242 & 68.27\\
NIPQ [$\lambda=6.0$] & 3.182 & 67.77\\
NIPQ [$\lambda=7.0$]& 3.162	& 67.69\\
NIPQ [$\lambda=10.0$]& 3.131 & 66.69\\
NIPQ [$\lambda=12.0$]&	\textbf{3.128} & \textbf{66.56}\\
\midrule
QBitOpt & \textbf{3.0} & \textbf{70.04}\\
\bottomrule
\end{tabular}
\caption{\efnet{Lite} on ImaneNet. NIPQ~\cite{Park2022NIPQ} results for different regularization strength $\lambda$, including our QBitOpt result. The values in bold are used in table~\ref{tbl:imagenet_results_avg_bits}.}\label{tbl:appedix_nipq_effnet}
\end{table}

In tables~\ref{tbl:appedix_NIPQ_mnv2} and~\ref{tbl:appedix_nipq_effnet}, we present the results of our hyperparameter search over the regularization strength $\lambda$ in the objective of \textit{noise injection pseudo quantization}  (NIPQ)~\cite{Park2022NIPQ}. Similar to DQ, we observe that we need an extensive grid search over the regularization parameter to find a satisfactory average \bwt solution. In fact, for NIPQ, we needed significantly stronger regularization values for all models.

\subsection{Effect of $\alpha$-factor in exponential moving average on sensitivity estimation}
\label{sec:ema_abl}
\begin{figure}[h]
    \centering
    \hspace{-3mm}\includegraphics[width=0.42\textwidth]{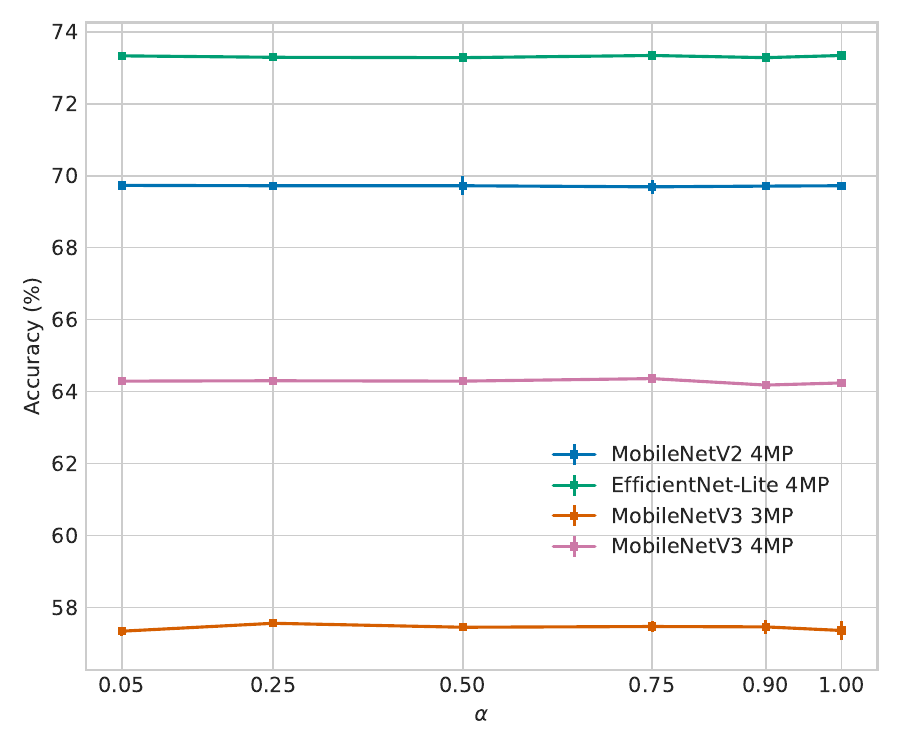}
    \caption{Accuracy obtained for different settings of $\alpha$ for
    the Exponential Moving Average sensitivity estimation. The whiskers, if visible, show the standard deviation over three seeds.}
    \label{fig:ema_results}
\end{figure}

In Section~\ref{sec:quant_sensitivities}, it was mentioned that an
exponential moving average (EMA) is used to estimate sensitivities over multiple batches during training. Specifically, given the observed sensitivity $e_t$ at timestep $t$ and $\alpha \in (0, 1]$, the EMA $\mathcal{E}_t$ is defined as:
\begin{align}
    \mathcal{E}_t = (1-\alpha)\mathcal{E}_{t-1} + \alpha e_t,\quad \mathcal{E}_1 = e_1.
\end{align}
The choice of $\alpha$ determines how much weight is put on past
observations compared to new observations. E.g., when $\alpha=1$, only the latest observations are used. Figure~\ref{fig:ema_results} shows results for different architectures trained using various values for $\alpha \in \{0.05, 0.25, 0.5, 0.75, 0.9, 1.0\}$. This shows that the choice of $\alpha$ is insignificant, and one could choose not to include the exponential moving average estimator. This result agrees with the progression of FIT estimates shown in Figure~\ref{fig:fit_sensitivities}, where we showed that sensitivities are relatively stable during training. However, since this figure also shows several outliers, we use the EMA estimator for the results
presented in the main text.

\subsection{QBitOpt bitwidth allocation}
\label{app:bitwidth_allocation}
\begin{figure*}[t]
     \centering
     \begin{subfigure}[b]{0.75\textwidth}
         \centering
         \includegraphics[width=1.\textwidth]{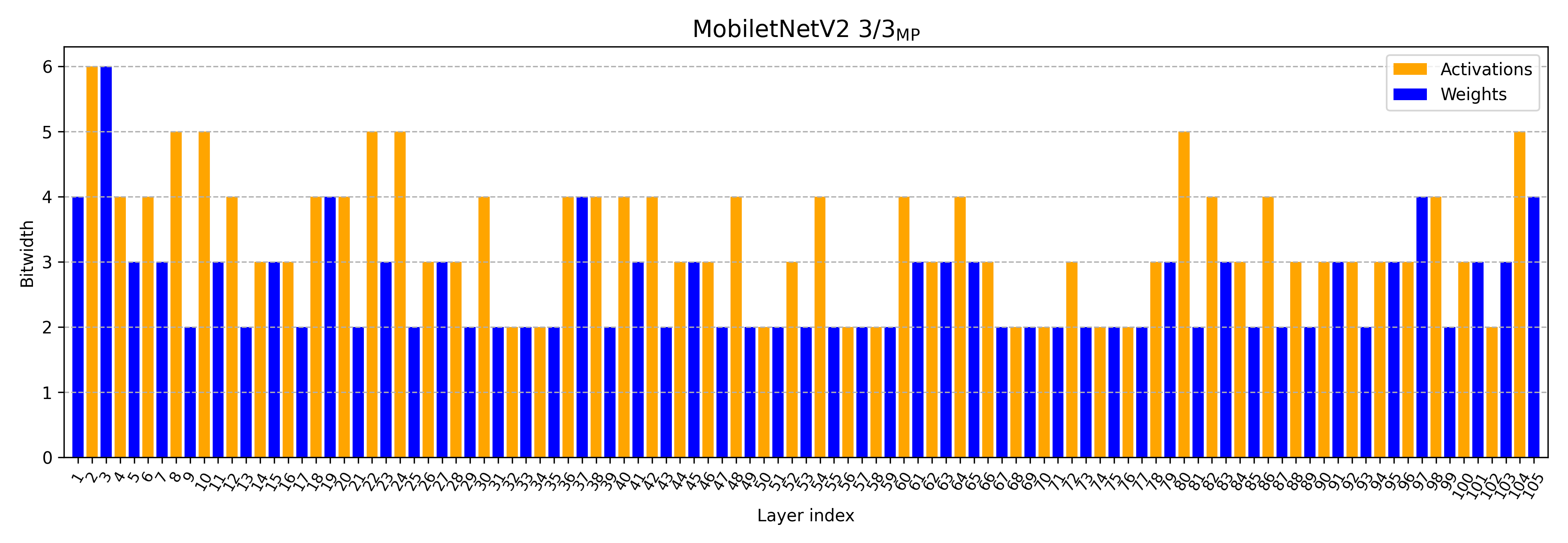}
\vspace{-.3cm}
         \label{fig:mnv2_w3a3_bits_allocation}
     \end{subfigure}
     \hfill
     \begin{subfigure}[b]{0.75\textwidth}
         \centering
         \includegraphics[width=1.0\textwidth]{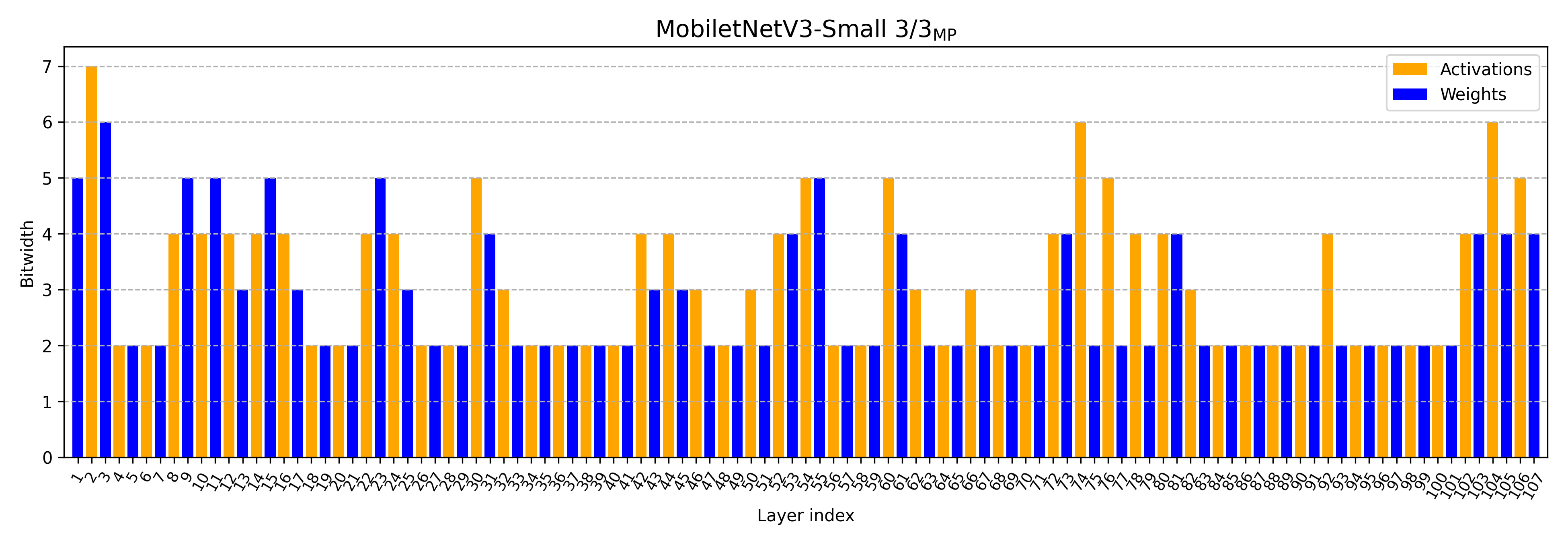}
        \label{fig:mnv3_w3a3_bits_allocation}
     \end{subfigure}
 \vspace{-.5cm}
    \caption{Final \bwt allocation for \mnv{2} and \mnv{3}-Small using QBitOpt with 3-bit average target \bwt \mpb{3/3}.}
        \label{fig:bits_allocation}
\end{figure*}
In figure~\ref{fig:bits_allocation}, we illustrate the final \bwt allocation resulting from QBitOpt on \mnv{2} and \mnv{3}-Small with a 3-bit average target bitwidth. In both cases, QBitOpt allocated  more bits in the first and last layers of the networks, which is consistent with empirical observations from existing literature. In addition, in \mnv{2}, we also observe that QbitOpt tends to assign higher \bwt to activations compared to weights.


\end{document}